\documentclass{pinepaper}

\usepackage{eccvabbrv}              
\usepackage{wrapfig}                
\usepackage{multirow}              
\usepackage{placeins}              
\usepackage{orcidlink}             
\usepackage[capitalize]{cleveref}  

\crefname{figure}{Fig.}{Figs.}
\Crefname{figure}{Figure}{Figures}
\crefname{table}{Tab.}{Tabs.}
\Crefname{table}{Table}{Tables}
\crefname{section}{Sec.}{Secs.}
\Crefname{section}{Section}{Sections}
\crefname{equation}{Eq.}{Eqs.}
\Crefname{equation}{Equation}{Equations}

\makeatletter
\renewcommand{\_}{\textunderscore\penalty\z@\hskip\z@}
\makeatother

\title{DVG-WM: Disentangled Video Generation Enables Efficient Embodied World Model for Robotic Manipulation}

\author[1]{Ziyu Shan\,\orcidlink{0000-0002-3346-4261}}
\author[2]{Zhenyu Wu\,\orcidlink{0009-0002-4827-6017}}
\author[3]{Xiaofeng Wang\,\orcidlink{0000-0002-0291-0230}}
\author[3]{Zheng Zhu\,\orcidlink{0000-0002-4435-1692}}
\author[1]{Ziwei Wang\textsuperscript{\textdagger}\,\orcidlink{0000-0001-9225-8495}}
\affil[1]{Nanyang Technological University, Singapore}
\affil[2]{Beijing University of Posts and Telecommunications, Beijing, China}
\affil[3]{GigaAI}

\date{\fontfamily{\pineauthorfamily}\selectfont\small \textsuperscript{\textdagger}Corresponding author.}

\projectpage{http://zyshan0929.github.io/DVGWM}
\correspondence{\href{mailto:ziwei.wang@ntu.edu.sg}{ziwei.wang@ntu.edu.sg}}

\begin{document}
\makepinetitle

\begin{pineabstract}
Video-based embodied world models provide an appealing substrate for robotic manipulation by predicting future states, yet current approaches remain limited by a fundamental entanglement: accurately modeling dynamics typically requires low-level temporal reasoning, while producing high-resolution frames demands expansive visual synthesis according to high-level semantics. This entanglement results in slow inference speed for iterative planning or too coarse predictions to retain contact-rich details. To solve this dilemma, we present Disentangled Video Generation World Model~(DVG-WM), an efficient framework that explicitly \textbf{decomposes world modeling into dynamics learning and visual synthesis}. Conditioned on an initial observation and a language instruction, our model first generates a plausible sequence of intermediate visual states to preview the physical interaction and refines them to obtain high-fidelity videos. Furthermore, an efficient cascading mechanism is proposed, where DVG-WM leverages flow matching to directly map the dynamics to video latents, and introduces a latent degradation mechanism to enable the capability of regenerating contact-rich details. Experiments on LIBERO and real-world platforms demonstrate improved video quality with up to 3.97 $\times$ acceleration, validating that disentangled video generation can be an efficient embodied world model for robotic manipulation.
\end{pineabstract}

\keywords{Robotic Manipulation, Embodied World Model, Imitation Learning}

\section{Introduction}
\label{sec:intro}

Embodied world models have gained increasing attention in the robotics community for their capability to model and predict the physical dynamics based on an initial observation and a conditioning input \cite{guo2025ctrl,zhang2025step,deng2025survey}. The action-conditioned world models support evaluating policies within imagination by simulating fine-grained action outcomes, while another significant role of embodied world models is to support goal-conditioned action planning by forecasting future states towards the given goal images or initial language instructions \cite{zhou2025act2goal,li2025keyworld,chen2026bridgev2w}.

\begin{figure*}[t]
  \centering
  \includegraphics[width=1\linewidth]{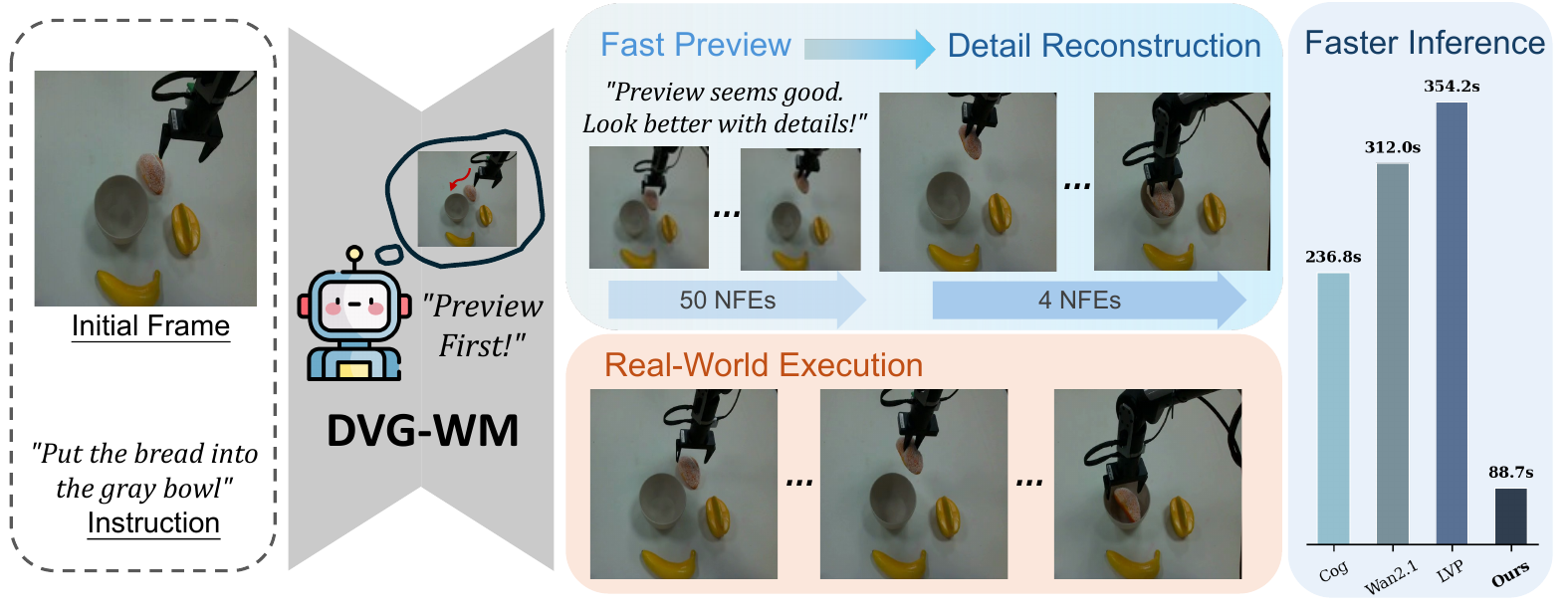}
  \caption{Illustration of our DVG-WM. Given the initial observation and a language instruction, DVG-WM first generates a sequence of low-resolution imagined future trajectory to preview the physical interaction, then reconstructs the high-fidelity video with faster inference speed (right). In addition to the improved visual quality, DVG-WM also achieves impressive performance on real-world manipulation tasks.}
    \vspace{-0.4cm}
  \label{fig:teaser}
\end{figure*}

The mainstream of embodied world models generally employs video prediction \cite{yang2024cogvideox,zhang2025flashvideo,wan2025wan} as the paradigm of world modeling. The typical pipeline involves fine-tuning a pretrained video generation backbone with conditioning input (\eg, language, action, and goal images) to produce latent dynamics or pixel-level future predictions \cite{chen2025large,zhou2025act2goal}. Despite the impressive visual quality and temporal coherence, the generated videos fail to satisfy the requirement of efficient inference for iterative planning in real-world scenarios. This inefficiency of these world models stems from a fundamental limitation: \textit{dynamics modeling is tightly entangled with high-fidelity visual synthesis}. Specifically, accurate dynamics modeling requires low-level temporal reasoning to produce smooth transitions such as contacts, slips, and occlusions, which concentrates on physical interaction patterns and is not necessarily learned in high-resolution settings with redundant textures \cite{xiao2025world,lu2025gwm}. Conversely, producing high-resolution frames demands expansive visual synthesis based on high-level semantics and rich texture information of the surroundings \cite{xie2025star}. Existing approaches \cite{zhu2025irasim,zhou2025act2goal,liao2025genie} often attempt to solve both requirements within a single monolithic video generator, resulting in slow inference speed for high-resolution observations or coarse predictions to retain contact-rich details. This becomes particularly problematic in real-world iterative planning where the world model must be queried repeatedly to predict the interaction process.


A straightforward solution to this dilemma is to learn the dynamics and visual synthesis separately. For the typical diffusion-based video world models \cite{zhou2025act2goal,li2025keyworld,shang2025longscape}, learning dynamics and visual synthesis simultaneously requires a large number of function evaluations (NFEs, \ie denoising steps for diffusion models \cite{ho2020denoising}) under high-resolution settings. In contrast, learning dynamics in a low-resolution latent space can significantly reduce the computation load for each NFE, allowing for low-cost preview to ensure the physical interaction is reasonable and consistent with the language instruction. Furthermore, the predicted latent dynamics can be refined into high-fidelity videos with fine-grained details with a minimal number of NFEs.
This disentangled design allows the world model to efficiently learn the dynamics and visual synthesis separately in their optimal resolution.

Following this philosophy, we propose Disentangled Video Generation World Model (DVG-WM), an efficient video-based world model for robotic manipulation that explicitly decomposes dynamics learning and visual synthesis, as shown in \cref{fig:teaser}. Conditioned on an initial observation and a language instruction, DVG-WM first generates a sequence of low-resolution intermediate visual states that captures the plausible evolution of the interaction, providing a fast dynamics preview suitable for iterative planning. 
Subsequently, the refinement stage produces high-fidelity videos with minimal NFEs (\ie 4 steps) to restore high-frequency details and contact-sensitive cues. Furthermore, a novel cascading mechanism is developed to efficiently connect the two stages, where flow matching is leveraged to directly map the low-resolution dynamics to high-resolution video latents, and a latent degradation mechanism is also introduced to enhance the model's capability to regenerate accurate structures of objects and the end-effector at high resolutions, which is crucial for contact-rich manipulation.

We evaluate DVG-WM on LIBERO \cite{liu2023libero} and real-world platforms, showing that disentangled generation yields high-quality visual predictions with up to $3.97\times$ speed up compared to the baselines. For the real-world tasks, the DVG-WM achieves better performance equipped with an action expert \cite{chi2025diffusion, ze20243d}, demonstrating its capability of serving as an action planner. Furthermore, the ablation studies demonstrate the disentangled design successfully learns the expected patterns separately. These results reveal that disentangled video generation can be a practical foundation for embodied world models.
The main contributions of this paper are summarized as follows:
\begin{itemize}
  \item We propose DVG-WM, a disentangled video generation world model that explicitly decomposes dynamics learning and visual synthesis into a two-stage pipeline, enabling efficient preview and refinement for robotic manipulation. 
  \item We propose a novel cascading mechanism to connect stages. Specifically, we leverage flow matching to directly map the dynamics to video latents, and introduce a latent degradation mechanism to enhance the model's capability to regenerate accurate structures for contact-rich manipulation.
  \item We evaluate DVG-WM on both simulation and real-world platforms, demonstrating improved visual prediction quality and accuracy with limited inference cost, validating disentangled video generation as an efficient embodied world model for robotic manipulation.
\end{itemize}

\section{Related Works}
\subsection{Embodied World Models}
According to applied condition types, the embodied world models can be categorized into two main types: action-conditioned and goal-conditioned models. 
The action-conditioned world models typically learn the outcomes of actions by predicting future latent states or videos, improving the policy generalization capability \cite{zhu2025irasim,agarwal2025cosmos,huang2025enerverse,shen2025videovla,guo2025ctrl,chi2025wow,bi2025motus,li2026causal}. Simultaneously, many works \cite{zhang2025imowm, huang2026pointworld, chen2026bridgev2w,lu2025gwm, shan2026dockanywhere, qian2025wristworld} focus on multi-modal generation conditioned on action, such as 3D point clouds, gaussians, depth, and mask. These action-conditioned world models can be incorporated into robotic manipulation as synthetic data generators that produce diverse successful rollouts for imitation learning and self-improvement \cite{chi2025diffusion,ze20243d,kim2026cosmos,ye2026world} in unseen scenarios, and serve as imagination-based evaluators that rank candidate policies \cite{team2025evaluating} or simulators for model-based reinforcement learning that reduce the sample complexity of real-world interaction \cite{wu2024ivideogpt,xiao2025world,zhu2025wmpo,jiang2025world4rl,yang2026rise}.

Another type of embodied world models is goal-conditioned, which learns to predict the future states towards a goal image or language instruction \cite{zhang2025chain,cen2025rynnvla}. These models can be used for planning by forecasting the future states and optionally output the action using an inverse dynamic model \cite{chen2025large}. Among these, KeyWorld \cite{li2025keyworld} and LongScape \cite{shang2025longscape} utilize language instruction as the goal and uses the world model as video planner. GE-Act \cite{liao2025genie} adopts a bi-model architecture with a world model predicting future visual features based on language instruction. WorldVLA \cite{cen2025worldvla} aligns vision and action learning within a unified latent space. Act2Goal \cite{zhou2025act2goal} introduces a visual goal image with multi-scale temporal sampling for closed-loop control. Despite the impressive visual quality of generated videos, the inference efficiency of these world models remains a bottleneck for real-world iterative planning. Our work addresses this issue by disentangling dynamics learning for the physical interactions and visual synthesis, enabling efficient preview and refinement for robotic manipulation.

\subsection{Video Diffusion}
Diffusion models \cite{ho2020denoising,chi2025diffusion,lipman2022flow} have emerged as the dominant framework for high-fidelity video synthesis \cite{ho2022imagen,blattmann2023stable,brooks2024video}. Video diffusion methods generate a sequence via iterative denoising of spatiotemporal representations, often in a compressed latent space to reduce memory and compute, while introducing explicit temporal modules to encourage coherence across frames. Beyond vanilla full-sequence denoising, recent work improves temporal modeling and long-horizon rollouts by imposing structured uncertainty over time, enabling causal or rolling generation beyond the training horizon \cite{ruhe2024rolling,chen2024diffusion}, and developing diffusion-based pipelines for long-horizon generation and super-resolution \cite{wang2025lavie,zhou2024upscale,xie2025star,zhang2025flashvideo,he2024venhancer}. These advances substantially improve temporal consistency and visual fidelity, but generally retain the high inference cost inherent to iterative denoising.
This limitation is particularly problematic for embodied manipulation, where iterative planning is needed. Our model addresses this mismatch by allocating computation asymmetrically through learning dynamics and visual synthesis at optimal resolutions. 

\section{Methodology}
\subsection{Problem Formulation}
Embodied world models aim to predict the dynamics of the environment by forecasting the future observations. Mathematically, given an initial observation image $x_0 \in \mathcal X$ and the condition $c$, the embodied world model learns a probabilistic generative model $p(x_{1:N}|x_0,c)$ to predict the future observations $x_{1:N}$ with a sequence length of $N$. 
Depending on the applications, the conditioning variable $c$ may take the form of actions to simulate the action-conditioned outcomes \cite{zhu2025irasim,guo2025ctrl}, or goals (\eg, language \cite{liao2025genie}, goal images\cite{zhou2025act2goal}) to plan the transition towards the expected future states. In this work, we focus on the latter setting conditioned on language instructions, enabling the world model to function as a language-guided planner for high-level task reasoning and interactive manipulation.
\begin{figure}[t]
  \centering
  \includegraphics[width=1\linewidth]{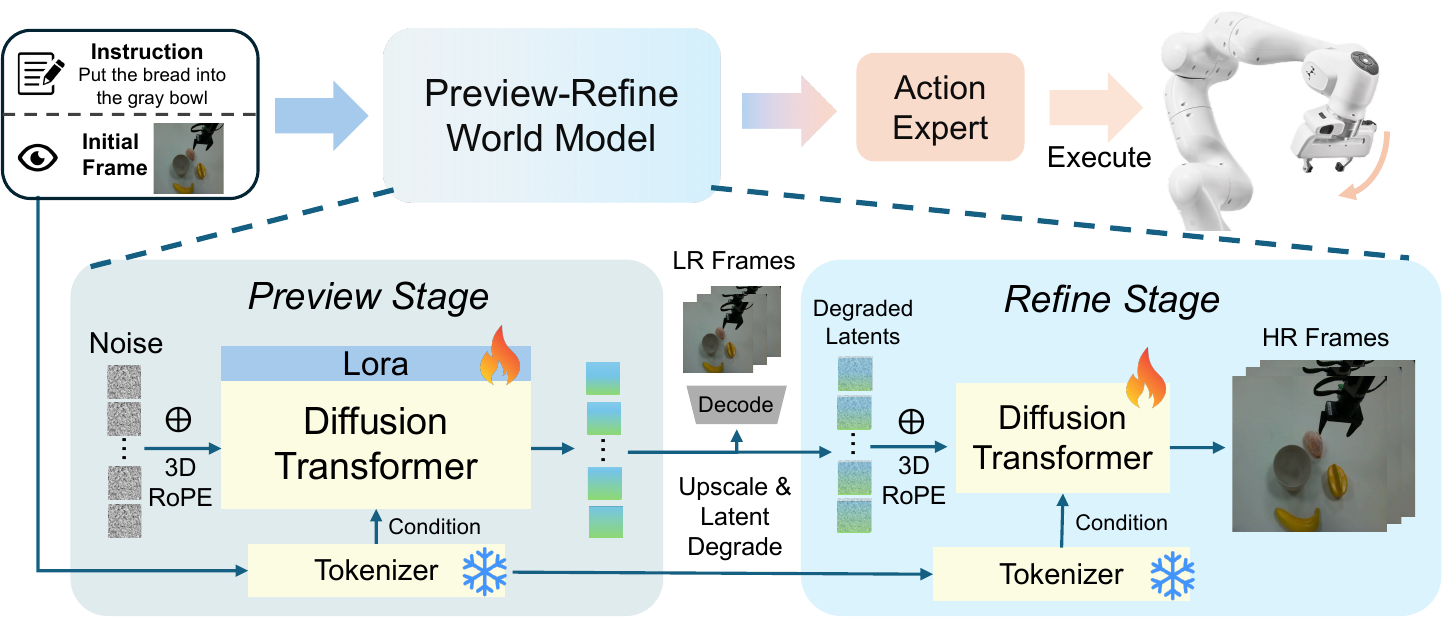}
  \caption{Pipeline of DVG-WM. The video world model consists of two main stages: 1) a preview stage that learns dynamics and generates low-resolution trajectories conditioned on language instructions and initial frames; 2) a refinement stage that upsamples the latent dynamics to produce high-fidelity video predictions with minimal additional computation. Finally, the inverse dynamics model extracts the action sequence from the predicted video for manipulation tasks.}
  \label{fig:pipeline}
\end{figure}

Existing video-based world models typically generate future frames $x_{1:N}$ at the original resolution (\eg, 720p, 1080p), which is computationally intensive due to the entanglement of dynamics modeling and high-fidelity visual synthesis. To enable efficient world modeling, we leverage 3D causal VAE \cite{yang2024cogvideox} to compress video pixels $x_T \in \mathbb{R}^{H\times W \times T}$ into latent features $z \in \mathbb{Q}^{h\times w \times t}$, where $h=H/8, w=W/8$ and $t=(T-1)/4+1$. 
Our model is designed to predict videos with 49 frames at 720p resolution. 
As shown in \cref{fig:pipeline}, we then propose a disentangled two-stage pipeline to learn the low-resolution (LR) dynamics and high-resolution (HR) visual synthesis separately, where each stage is optimized with tailored network architectures and training strategies for computational efficiency. The preview and refinement stages $g_{\text p}, g_{\text r}$ can be formulated as:
\begin{equation}
  g_{\text p}(z_{\text{lr}} | x_0^{\text d}, c); \quad g_{\text r}(z_{\text{hr}} | z_{\text{lr}}, x_0, c)
\end{equation}
where $x_0^{\text d}$ is the downsampled initial frame for the preview stage. The low-resolution dynamics can be optionally previewed by the 3D VAE decoder $\mathcal E$ as $x_{\text{lr}} = \mathcal E (z_{\text{lr}})$, while the high-resolution videos can be similarly obtained. The horizon $N$ of $z$ and $x$ is omitted for brevity.
The following sections provide detailed descriptions of each stage.

\subsection{Low-Resolution Preview Stage}
In the preview stage, the goal is to generate video latents $z_{\text{lr}}$ that align with the language instruction and captures the essential dynamics of the interaction. To achieve this, we adopt CogVideoX-5B \cite{yang2024cogvideox} as the video generation backbone. For computation efficiency, we discard the learnable positional embedding and employ flexible 3D rotary positional embedding \cite{su2024roformer} to allow for variable output resolutions. All other configurations such as language and image embedding, and denoising scheduler are the same as \cite{yang2024cogvideox}.
Then, we apply LoRA \cite{hu2022lora} with rank 128 on attention layers, FFN, and layer normalization to adapt the CogVideoX-5B to a lower resolution. 
Compared to full-parameter tuning, the LoRA-based adaptation significantly reduces the training cost while capturing the essential dynamic patterns, as demonstrated by the ablation studies in \cref{sec:ablation}. 
In this manner, the preview stage provides a physically meaningful initialization for the subsequent refinement stage, as detailed in the following section.

\subsection{High-Resolution Refinement Stage}
For the fine-grained visual synthesis, we employ a smaller model CogVideoX-2B due to the observation that embodied manipulation scenes are typically more structured, and their state transitions tend to be slower and more locally constrained compared to in-the-wild videos. As shown in \cref{fig:pipeline}, the refinement stage takes the low-resolution dynamics $z_{\text{lr}}$ as input and generates high-resolution video latents $z_{\text{hr}}$ with minimal NFEs.

\begin{wrapfigure}{R}{0.45\linewidth}
  \centering
  \includegraphics[width=\linewidth]{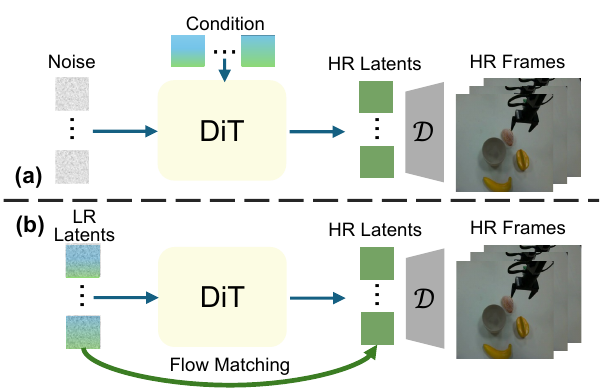}
  \caption{Comparison of refinement approaches. (a) Conventional DDPM treating low-resolution latents as conditioning input. (b) DVG-WM uses flow matching to directly map the low-resolution dynamics to high-resolution video latents.}
  \label{fig:compare}
\end{wrapfigure}
\noindent\textbf{Flow Matching for Cascaded Generation.} Given the low-resolution dynamics $z_{\text{lr}}$ as the input, the intuitive solution to generate high-resolution videos is to apply the diffusion process on another sequence of random noise conditioned on the dynamics. However, this approach consumes significant computation loads since the denoising process needs to learn to generate the entire video from scratch, which is redundant given that the preview stage already provides a reasonable initialization.

To solve this issue, we leverage flow matching \cite{liu2022flow,lipman2022flow} to directly establish the mapping from the low-resolution dynamics to the high-resolution video in the latent space. In our scenario, 
the flow matching learns the vector field $v_\theta\!\left(z_\tau,\tau \mid z_{\mathrm{lr}},x_0,c\right)$ that specifies a step-by-step update direction in latent space, where $z_\tau$ denotes an intermediate latent state at interpolation step $\tau\in[0,1]$, formulated as:
\begin{equation}
  z_\tau = (1-\tau)\,\tilde{z}_{\text{lr}} + \tau\,z_{\text{hr}}, \qquad \tau \sim \mathcal{U}(0,1).
  \label{eq:fm_path}
\end{equation}
where $\tilde{z}_{\text{lr}}$ is the linearly upsampled version of $z_{\text{lr}}$ to ensure identical sequence length.
In this manner, a straight-line probability path (\ie, an ODE trajectory for diffusion) is established, starting from the upscaled initialized latent $\tilde z_{\mathrm{lr}}$ and progressively following these directions to refine it into the final high-resolution latent $z_{\mathrm{hr}}$.

The path in \cref{eq:fm_path} linearly interpolates between the coarse initialization $\tilde{z}_{\mathrm{lr}}$ (at $\tau=0$) and the ground-truth latent $z_{\mathrm{hr}}$ (at $\tau=1$). 
As $\tau$ increases, the latent should move from $\tilde{z}_{\mathrm{lr}}$ toward $z_{\mathrm{hr}}$ along this straight line. 
Therefore, the desired update direction $u^\star(z_\tau,\tau)$ (\ie, velocity) at any intermediate point $z_\tau$ is simply the displacement from the start to the target:
\begin{equation}
u^\star(z_\tau,\tau) \triangleq \frac{d z_\tau}{d\tau} = z_{\mathrm{hr}} - \tilde{z}_{\mathrm{lr}}
\label{eq:fm_target_velocity}
\end{equation}

We train a neural vector field $v_\theta$ to predict this velocity, and the flow-matching objective can be formulated as:
\begin{equation}
  \mathcal{L}_{\text{FM}}
  = \mathbb{E}\Big[ \big\| v_\theta(z_\tau,\tau \mid z_{\text{lr}}, x_0, c)
  - u^\star(z_\tau,\tau) \big\|_2^2 \Big],
  \label{eq:fm_loss}
\end{equation}
where the expectation $\mathbb E$ is computed over the training samples $(\tilde{z}_{\text{lr}},z_{\text{hr}}, z_\tau, c)$,
and $\tau \sim \mathcal{U}(0,1)$.

At inference, refinement starts from the initialized latent $z^{0}=\tilde{z}_{\text{lr}}$ and applies limited NFEs through a deterministic flow process:
\begin{equation}
  z^{k+1} = z^{k} + \Delta\tau\; v_\theta\!\left(z^{k},\tau_k \mid z_{\text{lr}}, x_0, c\right),
  \qquad k=0,\dots,K-1,
  \label{eq:euler}
\end{equation}
where $\tau_k$ denotes the discretized interpolation step and $K$ is typically $4$ in our experiments, and the final output is $z_{\text{hr}} = z^{K}$ that can be decoded into high-resolution videos by the 3D VAE decoder $\mathcal D$.

\begin{figure}[t]
  \centering
  \includegraphics[width=1\linewidth]{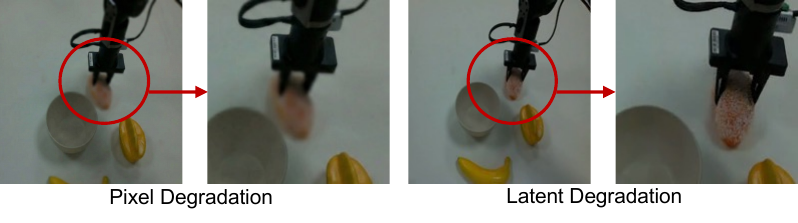}
  \vspace{-0.8cm}
  \caption{Comparison of pixel (left) and latent degradation (right) for training the refinement stage. Latent degradation generates contact-rich details in zoomed-in view.}
  \label{fig:degradation}
\end{figure}

\noindent\textbf{Latent Degradation for Contact-Rich Manipulation.} To train the refinement stage, the straightforward approach is to degrade the high-resolution video within the pixel level and encode it to form $z_\text{lr}$ so that the two stages can be trained independently, which is commonly used in video super-resolution \cite{xie2025star,zhou2024upscale}. 
However, this pixel-level degradation retains tight correlation between low- and high-resolution videos, constraining the world model from regenerating detailed structures for the interactions between the end-effector and objects. 
To address this issue, we use the latent degradation to perturb the upscaled output of the preview stage $\tilde{z}_{\text{lr}}$, allowing for \textbf{reconstructing fine-grained contact details }rather than upscaling pixels. More details can be found in the supplementary materials.
As shown in \cref{fig:degradation} of video prediction for a real-world task, the latent degradation successfully generates contact-rich details in the zoomed-in view, preserving the structures of the gripper and the target object, while the pixel degradation presents blur around the interaction region.

\subsection{Action Generation for Imitation Learning}
Given the high-resolution video prediction $x_{\text{hr}}$, we equip DVG-WM with an action expert to convert imagined futures into executable actions. We instantiate the action expert as a vision-only Diffusion Policy~(DP)~\cite{chi2025diffusion}, since robot states (\eg, joint angle) are unavailable for video prediction. 
The predicted frames of different timestamps are separately encoded as the condition $h_t$ of DP, which denoises a noise sequence into actions through the standard diffusion objective:
\begin{equation}
  \mathcal{L}_{\text{DP}} =
  \mathbb{E}\Big[\big\|\epsilon - \epsilon_\phi(a_\sigma, \sigma \mid h_t)\big\|_2^2\Big],
  \label{eq:dp_loss}
\end{equation}
where $a_\sigma$ is the noised action sequence at diffusion level $\sigma$ and $\epsilon \sim \mathcal{N}(0,I)$. Note that the action expert can also use diffusion-based models \cite{zhou2025act2goal,liao2025genie} and vision-language-action (VLA) models like NORA \cite{hung2025nora} that do not rely on robot state, which is unavailable for predicted future frames.

\noindent\textbf{Training Strategy.} We train the full system in two stages for stability. We first train the world model alone using the objectives in \cref{eq:fm_loss}, obtaining a reliable video predictor. Afterwards, we attach the action expert and optimize it on demonstrations while jointly fine-tuning the world model using only $\mathcal{L}_{\text{DP}}$. The gradients from the imitation learning loss propagate through both the action generation expert and the world model, aligning the visual prediction with downstream action planning.

\begin{table}[t]
\centering
\small
\setlength{\tabcolsep}{8pt}
\renewcommand{\arraystretch}{1.2}
\caption{Quantitative comparison on video quality and object-level accuracy.}
\begin{tabular}{lccccc}
\toprule
\textbf{Method} & \textbf{PSNR} $\uparrow$  & \textbf{SSIM} $\uparrow$ & \textbf{LPIPS} $\downarrow$ & \textbf{FVD} $\downarrow$ & \textbf{\shortstack{Object}} $\uparrow$\\
\midrule
CogVideoX-5B     & 19.286 & 0.761 & 0.138 & 171.24 & 76\% \\
Wan2.1-14B       & 18.964 & 0.732 & 0.162 & 198.54 & 68\% \\
LongScape        & 19.977 & \textbf{0.788} & 0.123 & 153.72 & - \\
LVP-14B              & 19.582 & 0.765 & 0.134 & 187.69 & 80\% \\
\rowcolor{gray!20}
DVG-WM (ours)    & \textbf{20.019} & 0.783 & \textbf{0.120} & \textbf{152.36} & \textbf{89\%} \\
\bottomrule
\end{tabular}
\vspace{-0.4cm}
\label{tab:quant_main}
\end{table}

\section{Experiments}
\subsection{Experimental Settings}
\noindent\textbf{Dataset Preparations.} The proposed DVG-WM is trained and evaluated on the simulation LIBERO dataset \cite{liu2023libero} and a self-collected real-world dataset containing 7K trajectories, aligning with the real-world platform setup in \cref{sec:real}. 
We evaluate the DVG-WM for video quality on LIBERO, while the real-world dataset is used to evaluate the performance of DVG-WM as an action planner.
For comprehensive evaluation with varied instructions, we split the official demonstrations according to the tasks with a ratio of 8:2 for training and testing. Additionally, we preprocess the dataset by replaying trajectories, removing dummy episodes and resizing, following \cite{cen2025worldvla,li2025keyworld}, obtaining around 5K trajectories in total. As for the real-world dataset, the trajectories with less than 49 frames are discarded. More details about the dataset are in supplementary materials.

\noindent\textbf{Implementation Details.} All videos are resized to $256\times 384$ for the preview stage and $480\times 720$ for refinement. For the preview stage, the model is trained using LoRA \cite{hu2022lora} for 10,000 iterations with a batch size of 4 on a single A100-80G GPU. AdamW optimizer is used with $\beta_1=0.9, \beta_2=0.95$, a weight decay of $1e-4$, gradient clipping setting to 0.1. 
For the refinement stage, the model is trained for 10 epochs with a batch size of 6 on LIBERO, taking around 24 hours on 8 A100-80G GPUs. The training hyper-parameters are kept the same as \cite{yang2024cogvideox}. More training details are in \cref{sec:real-world-task} and supplementary materials.

\noindent\textbf{Baselines.} We compare our approach with advanced video generation world models, including CogVideoX-5B \cite{yang2024cogvideox}, Wan2.1-14B I2V \cite{wan2025wan}, LongScape \cite{shang2025longscape}, and large video planner (LVP) \cite{chen2025large}. These models are fine-tuned on the same training set. Note that Wan2.1 and LVP are based on different resolutions, so an additional training set with the same trajectories is prepared.

\noindent\textbf{Evaluation Metrics.} For video quality evaluation, we use the standard metrics including PSNR \cite{zhang2024asynchronous,shan2024contrastive}, SSIM \cite{wang2004image}, LPIPS \cite{zhang2018unreasonable} and FVD. We also use a metric of object-level accuracy following \cite{li2025keyworld} where 10\% of trajectories are manually checked to report the ratio of the robot operating on the proper object. 
For the real-world tasks in \cref{sec:real}, we additionally evaluate two metrics to test the world model's understanding of the physical interaction process: 1) Mask-IoU between embodiment regions \cite{chen2026bridgev2w} in generated and ground-truth frames, obtained via Grounded SAM \cite{ren2024grounded} with “robotic arm” as the textual prompt. 2) Success rates of the executed actions planned by DVG-WM equipped with an action expert on nine real-world tasks across two platforms.

\begin{figure*}[t]
  \centering
  \includegraphics[width=1\linewidth]{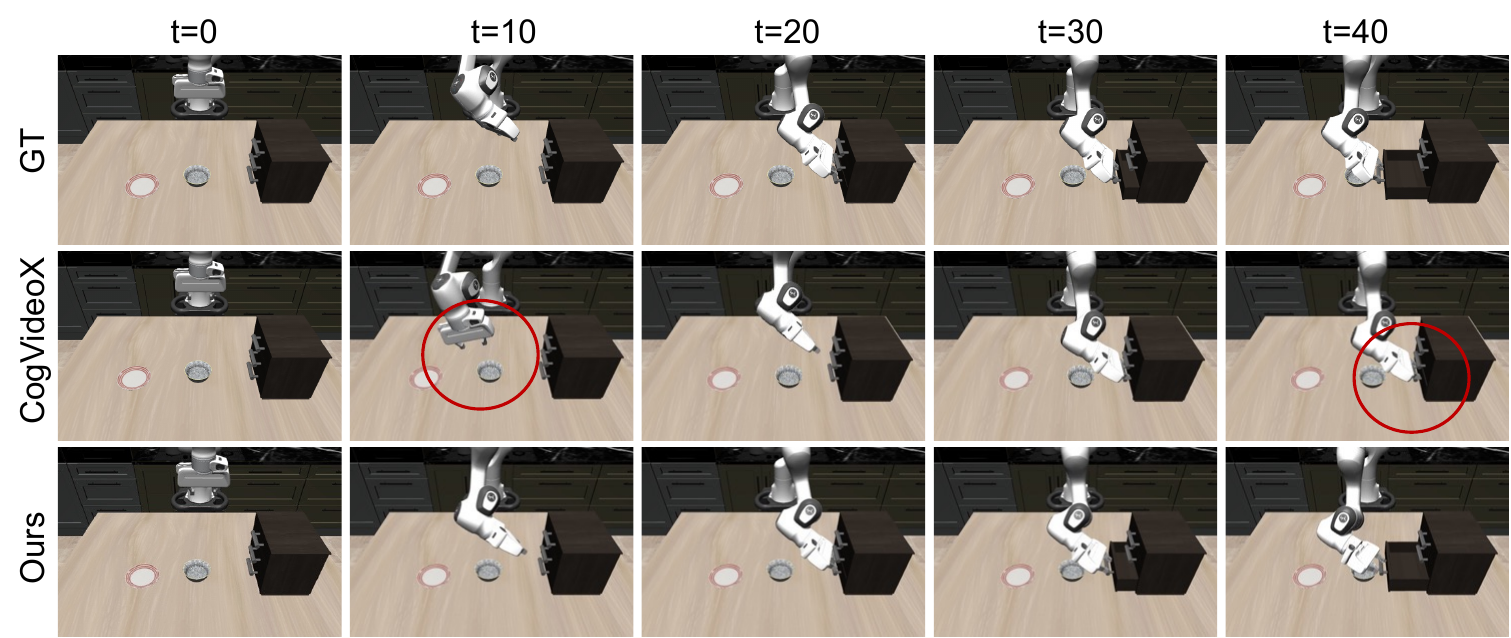}
  \caption{Qualitative comparison of video predictions from DVG-WM and CogVideoX on LIBERO tasks. Red circles highlight wrong critical interaction details of CogVideoX.}
  \label{fig:qualitative}
\end{figure*}

\subsection{Main Results}

\noindent\textbf{Quantitative Comparison.} In \cref{tab:quant_main}, our DVG-WM delivers stronger video prediction on LIBERO. Our method achieves superior visual quality metrics with an average improvement of 3.7\%, while showcasing impressive efficiency. 
Although SSIM is slightly lower than LongScape, it remains competitive, which reveals that the refinement stage improves high-frequency details without introducing severe structural distortion. 
Note that the original metrics of LongScape are reported and object-level accuracy is inaccessible.
Beyond low-level similarity, DVG-WM substantially improves object-level accuracy to 89\%, showing that the model more reliably grounds the interaction on the correct target objects, which is critical for manipulation-centric tasks. 
Achieving both high fidelity and reliable instruction following underscores the benefit of our two-stage design, where the preview supplies contact-rich cues that steer refinement towards physically meaningful, high-quality generation.

\noindent\textbf{Qualitative Comparison.} As shown in \cref{fig:qualitative}, CogVideoX is compared with our method since they share the same resolution. Obviously, our DVG-WM generates high-quality videos, and forecasts the complete trajectory for the open-drawer task, compared to CogVideoX. Despite the satisfactory visual quality, CogVideoX fails to accurately follow the instruction, where the robot arm shows the intention to interact with the wrong object (the bowl) instead of the drawer at $t=10$. Additionally, the final interaction stage of pulling the drawer is missing at $t=40$ in CogVideoX, which is critical for the task completion. 
In contrast, DVG-WM successfully captures the contact-rich details of the interaction, such as the gripper's contact with the drawer and the subsequent movement of the drawer, demonstrating its superior capability to model the physical dynamics and follow instructions accurately.

\begin{wraptable}{r}{0.4\linewidth}
  \centering
  \small
  \setlength{\tabcolsep}{1pt}
  \renewcommand{\arraystretch}{1}
  \caption{Comparison of inference time for the video world models.}
  \begin{tabular}{lc}
  \toprule
  \textbf{Method} & \textbf{Inference time} \\
  \midrule
  CogVideoX-5B & 236.8s \\
  Wan2.1-14B & 312.0s \\
  LVP-14B & 354.2s \\
  \rowcolor{gray!20}
  DVG-WM (ours) & \textbf{88.7s} \\
  \bottomrule
  \end{tabular}
\end{wraptable}

\noindent\textbf{Efficiency Comparison.} Our DVG-WM achieves a significant speedup of up to $3.97\times$ compared to LVP, since only 4 denoising steps is needed for refinement and the 50 denoising steps are conducted at the much lower resolution, which is crucial for real-world iterative planning. 
The efficiency can be attributed to the disentangled design that learns dynamics and visual synthesis separately in their optimal resolution settings.

\subsection{Ablation Studies}
\label{sec:ablation}

\begin{figure}[t]
  \centering
  \includegraphics[width=1\linewidth]{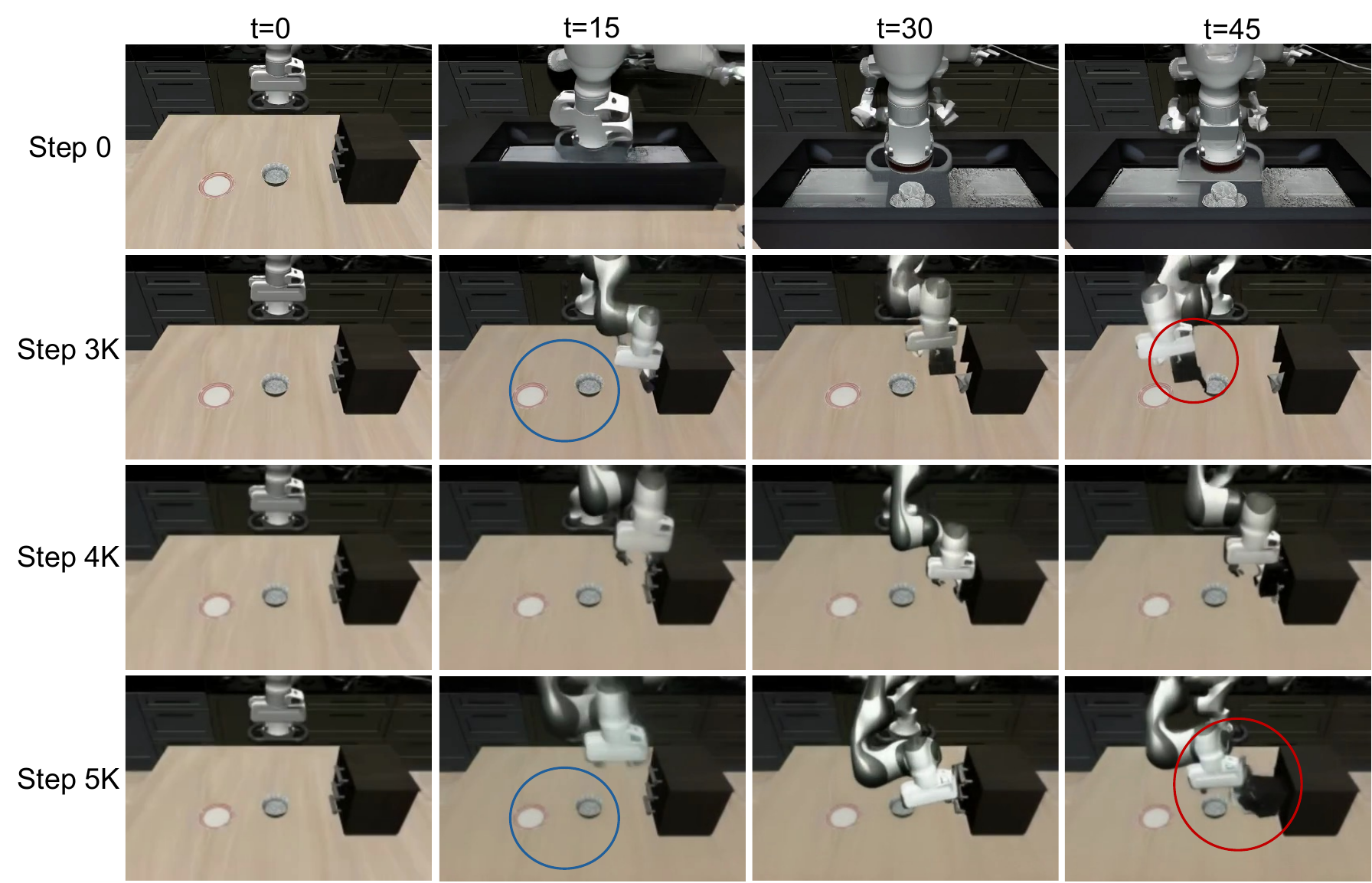}
  \caption{Comparison with generated videos of preview stage with different LoRA fine-tuning steps. Blue circles highlight the texture of the bowls, while the red circles highlight the contact details between the gripper and the drawer.}
  \label{fig:ablation}
\end{figure}

\noindent\textbf{Diving into the Disentangled Design.} To further validate the effectiveness of the disentangled model design, we conduct a qualitative comparison to analyze the effect of the LoRA fine-tuning for the preview stage in \cref{fig:ablation}.
For step 0, as well as the zero-shot inference for CogVideoX-5B, the layouts of the scene are totally changed and the robot arm is missing, even though the texture of the scene is fine-grained. This is expected since the original model is pretrained on in-the-wild videos, not including embodied data.

As the LoRA iteration step increases, the preview stage fails to restore the original textures of the scene, including the objects and robot arms, as highlighted in blue circles. This blur is even be aggravated as the iteration step increases. This loss of texture synthesis is expected since the preview stage is optimized for dynamics modeling rather than visual synthesis. 
However, the preview stage successfully captures the contact-rich details of the interaction, such as the gripper's contact with the drawer and the subsequent movement of the drawer, as highlighted in red circles. 
For step 3K, the drawer is unreasonably pulled out, while the model at step 5K successfully models this interaction, revealing the understanding of the instruction and interaction process. 
This observation validates that the preview stage learns to capture the essential dynamic patterns for physical interactions, which provides a meaningful initialization for the refinement stage.

\begin{wrapfigure}{R}{0.47\linewidth}
  \centering
  \includegraphics[width=\linewidth]{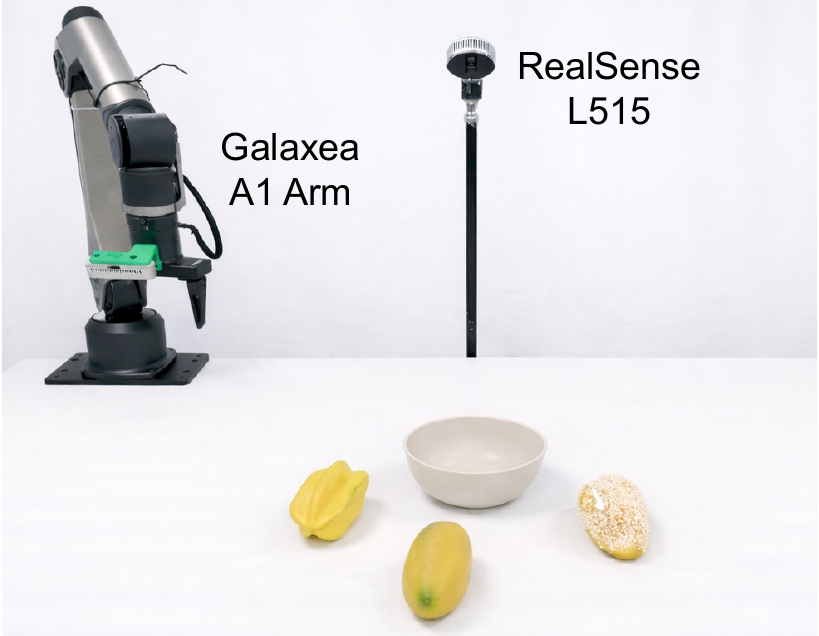}
  \caption{Real-world setup for Galaxea A1 Arm Platform.}
  \label{fig:setup}
\end{wrapfigure}
Meanwhile, the CogVideoX that is used for the refinement stage successfully restores the texture details of the scene, as in the second row of \cref{fig:qualitative}, but fails to plan the correct trajectory. Combined with the dynamics preview, the full model successfully generates high-fidelity videos with accurate interaction dynamics, demonstrating the effectiveness of the disentangled design, as in the last row of \cref{fig:qualitative}.


\begin{table}[t]
\centering
\small
\setlength{\tabcolsep}{6pt}
\renewcommand{\arraystretch}{1.1}
\caption{Component analysis of our DVG-WM on LIBERO.}
\begin{tabular}{c l cccc}
\toprule
\textbf{\#Stages} & \textbf{Method} & \textbf{PSNR} $\uparrow$ & \textbf{SSIM} $\uparrow$ & \textbf{LPIPS}$\downarrow$ & \textbf{FVD} $\downarrow$ \\
\midrule
\multirow{4}{*}{Two} 
& \cellcolor{gray!20}DVG-WM (Ours) 
& \cellcolor{gray!20}\textbf{20.019} 
& \cellcolor{gray!20}\textbf{0.783} 
& \cellcolor{gray!20}\textbf{0.120} 
& \cellcolor{gray!20}\textbf{152.36} \\
& Pixel Degradation & 19.421 & 0.768 & 0.128 & 168.54 \\
& Naive Cascading & 19.580 & 0.765 & 0.134 & 187.69 \\
& 360p Preview + Refine     & 18.812 & 0.756 & 0.146 & 188.47 \\
\midrule
\multirow{2}{*}{One}
& Only Preview    & 14.714 & 0.721 & 0.341 & 238.72 \\
& Only Refine       & 19.277 & 0.760 & 0.135 & 169.29 \\
\bottomrule
\end{tabular}
\vspace{-0.4cm}
\label{tab:ablation}
\end{table}

\noindent\textbf{Component Analysis.} We conduct component ablations on LIBERO to validate the design choices of DVG-WM in \cref{tab:ablation}. We have multiple observations: 
1) Replacing latent degradation with pixel-level degradation consistently degrades quality (SSIM 0.783$\rightarrow$0.768), supporting our intuition that pixel-space pairing causes tight correlation between the source and target, leading to shortcut upscaling. Conversely, latent degradation reconstructs contact-relevant structures with much higher visual and structural quality.
2) Naive cascading that treats the low-resolution dynamics as conditioning input for the refinement stage (in \cref{fig:compare} (a)) leads to a performance drop, validateing the effectiveness of flow matching in directly mapping the dynamics to high-resolution video latents, rather than generating from random noise. 3) The preview and refinement stages

 are both necessary. Removing the preview stage reduces all metrics, confirming that the low-resolution latents provide informative cues that guide refinement. Similarly, removing the refinement stage leads to a severe collapse in perceptual quality, showing that dynamics-only low-resolution predictions are insufficient to preserve fine-grained details. 
4) Lowering the preview resolution to $360 \times 240$ further worsens the performance, which may be attributed to the failure to leverage the 
pretraining knowledge of CogVideoX that is large-scale pretrained on $256\times 384$ videos.

\begin{wraptable}{r}{0.42\linewidth}
  \centering
  \scriptsize
  \setlength{\tabcolsep}{2pt}
  \renewcommand{\arraystretch}{1.0}
  \caption{Ablation over the number of refinement steps $K$ on LIBERO.}
  \begin{tabular}{ccccc}
  \toprule
  $K$ & \textbf{PSNR} $\uparrow$ & \textbf{SSIM} $\uparrow$ & \textbf{LPIPS} $\downarrow$ & \textbf{FVD} $\downarrow$ \\
  \midrule
  1 & 16.514 & 0.681 & 0.218 & 245.83 \\
  2 & 18.702 & 0.745 & 0.165 & 198.41 \\
  \rowcolor{gray!20}
  4 & \textbf{20.019} & \textbf{0.783} & \textbf{0.120} & \textbf{152.36} \\
  8 & 20.124 & 0.785 & 0.118 & 150.42 \\
  \bottomrule
  \end{tabular}
  \label{tab:refine_steps}
\end{wraptable}

\noindent\textbf{Number of Refinement Steps.} We further study the effect of the number of refinement steps $K$ in the refinement stage on LIBERO, as reported in \cref{tab:refine_steps}. Increasing $K$ from $1$ to $4$ consistently improves all metrics, since additional denoising steps progressively recover fine-grained textures from the low-resolution dynamics. However, doubling $K$ from $4$ to $8$ yields only marginal gains (\eg, FVD $152.36{\rightarrow}150.42$) while nearly doubling the refinement computation. We therefore adopt $K{=}4$ by default, striking a favorable balance between visual fidelity and inference efficiency.
\subsection{Real-World Experiments}
\label{sec:real}

\noindent\textbf{Platform Setup.} We conduct real-world experiments on two robot platforms, a Galaxea
A1 robotic arm and a UR7e robotic arm, each equipped with a parallel gripper. The observation is provided by a RealSense L515 RGBD camera, as shown in \cref{fig:setup}.
The layouts are aligned with the real-world dataset used for training the world model.

\label{sec:real-world-task}
\noindent\textbf{Real-World Tasks and Results.} We evaluate DVG-WM on nine real-world manipulation tasks spanning two robot platforms (Galaxea A1 and UR7e): move banana, pick-and-place bread, sweep rubbish, stack cube, adapter insertion, push cube, open drawer, assemble gears, and make a toast. For each task, around 50 demonstrations are prepared for imitation learning in \cref{eq:dp_loss}. To evaluate the models, we compare the video quality sharing the protocol in \cref{tab:quant_main}, the mask-IoU for the embodiment regions between the generated and ground-truth videos, and the actual success rates for the tasks. The dataset is resized to $480 \times 720$ to train the CogVideoX and our DVG-WM, along with the action expert of DP, whose input horizon is set to 1 to adapt to the iterative planning. We roll out the learned policy 20 times for each task under four distinct settings with various tabletop layouts, instance locations, and task horizons, yielding $20\times4{=}80$ rollouts per task, and report the average success rate with standard deviation.

\begin{figure}[t]
  \centering
  \includegraphics[width=1\linewidth]{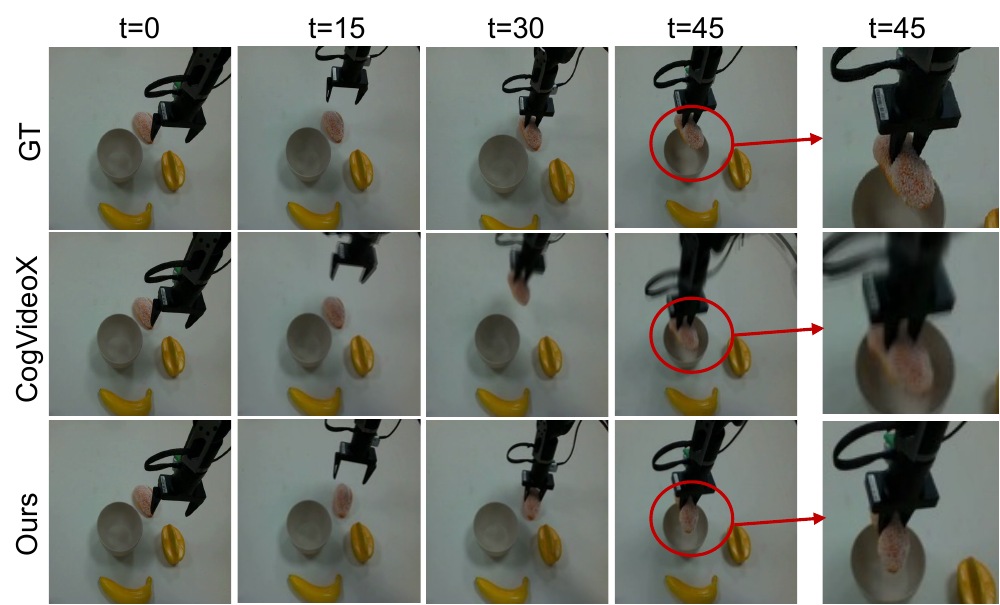}
  \caption{Qualitative results on the real-world manipulation task of placing bread into bowl. DVG-WM generates high-fidelity video predictions that accurately capture the interaction details and follow language instructions for pick-and-place.}
  \label{fig:real-qualitative}
\end{figure}

\begin{table}[t]
\centering
\setlength{\tabcolsep}{6pt}
\renewcommand{\arraystretch}{1.2}
\caption{Quantitative comparison on video quality for real-world scenarios.}
\begin{tabular}{lccccc}
\toprule
\textbf{Method} & \textbf{PSNR} $\uparrow$  & \textbf{SSIM} $\uparrow$ & \textbf{LPIPS} $\downarrow$ & \textbf{FVD} $\downarrow$ & \textbf{Mask-IoU} $\uparrow$\\
\midrule
CogVideoX-5B     & 18.642 & 0.745 & 0.151 & 189.32 & 0.54 \\
\rowcolor{gray!20}
DVG-WM (ours)    & \textbf{19.287} & \textbf{0.771} & \textbf{0.128} & \textbf{164.78} & \textbf{0.57} \\
\bottomrule
\end{tabular}
\label{tab:real_world}
\end{table}

As shown in \cref{tab:real_world}, DVG-WM outperforms CogVideoX-5B across all video quality metrics, demonstrating its superior capability to generate high-fidelity videos that accurately capture the dynamics of real-world interactions. The improvement in Mask-IoU indicates that DVG-WM better perceives the embodiment regions, which is crucial for understanding the physical interactions between the robot and objects.

For visualization, DVG-WM successfully predicts realistic future interactions on all two tasks in \cref{fig:real-qualitative}, demonstrating its capability to generalize from simulation to real-world scenarios. The predicted videos of CogVideoX present slight ghosting artifacts and motion blur, as revealed in \cite{shang2025longscape}. Conversely, our DVG-WM exhibits fine-grained contact details between the gripper and objects and proper task completion sequences, even the contact area and texture of the bread preserved, demonstrating its capability to be utilized as a real-world video planner.

\begin{table}[t]
\centering
\setlength{\tabcolsep}{4pt}
\renewcommand{\arraystretch}{1.0}
\caption{Success rates (\%) on nine real-world tasks across two platforms (Galaxea A1 \& UR7e), each evaluated with $20\times4{=}80$ rollouts. \textbf{Avg.}~is the mean over all nine tasks.}
\label{tab:sr}
\vspace{-0.15cm}
\resizebox{0.72\textwidth}{!}{%
\begin{tabular}{lccccc}
\toprule
\textbf{Method} & \shortstack{move\\banana} & \shortstack{pick\&place\\bread} & \shortstack{sweep\\rubbish} & \shortstack{stack\\cube} & \shortstack{adapter\\insertion} \\
\midrule
CogVideoX & $48.4{\pm}11.6$ & $43.2{\pm}6.8$ & $38.6{\pm}11.4$ & $26.4{\pm}8.6$ & $18.2{\pm}6.8$ \\
\rowcolor{gray!20}
DVG-WM (ours) & $\mathbf{73.2{\pm}12.8}$ & $\mathbf{67.8{\pm}7.2}$ & $\mathbf{72.6{\pm}7.6}$ & $\mathbf{57.8{\pm}12.2}$ & $\mathbf{52.4{\pm}7.6}$ \\
\bottomrule
\end{tabular}%
}

\vspace{0.2em}

\resizebox{0.72\textwidth}{!}{%
\begin{tabular}{lccccc}
\toprule
\textbf{Method} & \shortstack{push\\cube} & \shortstack{open\\drawer} & \shortstack{assemble\\gears} & \shortstack{make\\a~toast} & \textbf{Avg.} \\
\midrule
CogVideoX & $52.6{\pm}7.4$ & $41.4{\pm}8.6$ & $22.6{\pm}7.4$ & $14.4{\pm}5.6$ & $34.0$ \\
\rowcolor{gray!20}
DVG-WM (ours) & $\mathbf{78.4{\pm}6.6}$ & $\mathbf{68.2{\pm}11.8}$ & $\mathbf{51.4{\pm}8.6}$ & $\mathbf{38.2{\pm}11.8}$ & $\mathbf{62.2}$ \\
\bottomrule
\end{tabular}%
}
\vspace{-0.2cm}
\end{table}

To verify if the DVG-WM can be used for practical manipulation tasks, we equipped it with a vision-only DP as the action expert and jointly train them through imitation learning based on the provided demonstrations of the dataset. The same protocol is applied to every task, and the average success rates with standard deviation are reported in \cref{tab:sr}.
We observe that DVG-WM achieves an average success rate of 62.2\% across the nine tasks, outperforming the CogVideoX world model (34.0\%) by 28.2\% on average. The margin is most pronounced on contact-rich and long-horizon tasks such as adapter insertion, assemble gears, and make a toast, where accurate dynamics prediction is critical for completion. These consistent gains across diverse tasks and two platforms validate its effectiveness as an embodied world model for real-world robotic manipulation.
The demos and more details of the real-world experiments are in supplementary materials.
\section{Conclusion}
In this paper, we identify the bottleneck of inefficient video-based embodied world models: modeling dynamics and synthesizing high-resolution observations are tightly entangled.
We then present DVG-WM, a disentangled video generation world model that addresses this limitation by decomposing world modeling into a preview stage that captures dynamics and a high-resolution refinement stage restoring fine-grained details with minimal computation. To efficiently connect the two stages, we use flow matching to directly map the dynamics to video latents, together with a latent degradation mechanism that encourages reconstructing contact-rich structures rather than performing shortcut upscaling.
Experiments on LIBERO and real-world platforms demonstrate that DVG-WM improves video prediction quality and instruction-following accuracy while reducing inference time with up to 3.97$\times$ acceleration. When equipped with an action expert, DVG-WM further achieves higher real-world success rates with 28.2\% improvement. We believe disentangled video generation provides a promising direction for scalable embodied world models.
%
%
\section*{Acknowledgments}
This work was supported by MoE AcRF
Tier 2 under Grant MOE-T2EP50125-0004, and Singapore National Robotics Programme Research Project DS-RFM M25N4N2009, DEM M25N4N2150, MoE Tier 1 Seed Project RS17/24. The corresponding author is Ziwei Wang.

\bibliographystyle{splncs04}
\bibliography{main}

\clearpage
\appendix
\counterwithin{figure}{section}
\counterwithin{table}{section}
\counterwithin{equation}{section}
\begin{center}
  {\sffamily\bfseries\LARGE\color{PineInk} Appendix}
\end{center}
\vspace{0.6em}

\section{Details of Latent Degradation}
\label{app:latent_degradation}

The refinement stage aims to reconstruct a high-resolution latent video
$z_{\mathrm{hr}} \in \mathbb{R}^{c \times t \times h_{\mathrm{hr}} \times w_{\mathrm{hr}}}$ from the low-resolution dynamics produced by the preview stage. A straightforward way to construct training pairs is to first downsample the ground-truth video in pixel space and then re-encode it into a low-resolution latent. However, such a formulation makes the refinement problem overly close to a deterministic super-resolution task, since the degraded input remains strongly correlated with the target. As a result, the model tends to learn shortcut upscaling rather than regenerating fine-grained structures that are crucial for contact-rich manipulation.

To alleviate this issue, we simulate the refinement input by \emph{latent degradation}.
Let $x_{\mathrm{hr}} \in \mathbb{R}^{H \times W \times T}$ denote the ground-truth high-resolution video,
and let $\mathcal{E}(\cdot)$ be the 3D causal VAE encoder.
We first apply a pixel-space degradation operator
$\mathrm{Deg}_{\mathrm{pix}}(\cdot)$ to mimic the loss of high-frequency visual details:
\begin{equation}
\hat{x}_{\mathrm{lr}} = \mathrm{Deg}_{\mathrm{pix}}(x_{\mathrm{hr}}),
\end{equation}
where $\mathrm{Deg}_{\mathrm{pix}}(\cdot)$ includes spatial downsampling followed by resizing back
to the target resolution.\footnote{In practice, other blur or interpolation operators can also be used.}
The degraded video is then encoded into latent space:
\begin{equation}
z_{\mathrm{deg}} = \mathcal{E}(\hat{x}_{\mathrm{lr}}).
\label{eq:pixel_deg_latent}
\end{equation}

Although $z_{\mathrm{deg}}$ removes part of the texture information,
it is still tightly aligned with the target latent
$z_{\mathrm{hr}} = \mathcal{E}(x_{\mathrm{hr}})$.
Therefore, we further apply a latent degradation operator
$\mathrm{Deg}_{\mathrm{lat}}(\cdot)$ by perturbing the degraded latent with Gaussian noise:
\begin{equation}
\tilde{z}_{\mathrm{lr}}
= \mathrm{Deg}_{\mathrm{lat}}(z_{\mathrm{deg}}; s)
= \alpha_s \, z_{\mathrm{deg}} + \beta_s \, \epsilon,
\qquad
\epsilon \sim \mathcal{N}(0, I),
\label{eq:latent_deg}
\end{equation}
where $s$ denotes the degradation strength, and the coefficients
$\alpha_s, \beta_s \in \mathbb{R}$ satisfy
\begin{equation}
\alpha_s^2 + \beta_s^2 = 1.
\label{eq:alpha_beta_constraint}
\end{equation}
This parameterization preserves the overall latent scale while controlling how far
the refinement input deviates from the pixel-degraded latent.
A larger $\beta_s$ yields stronger perturbation and encourages the refinement model
to rely more on its learned generative prior to reconstruct plausible structures.

The resulting latent $\tilde{z}_{\mathrm{lr}}$ serves as the starting point of the refinement stage
during training. Concretely, instead of directly using a purely resized latent from the first stage,
we train the flow-matching refiner to transport
$\tilde{z}_{\mathrm{lr}}$ toward the target high-resolution latent $z_{\mathrm{hr}}$.
The interpolation path is defined as
\begin{equation}
z_{\tau} = (1-\tau)\,\tilde{z}_{\mathrm{lr}} + \tau\, z_{\mathrm{hr}},
\qquad \tau \sim \mathcal{U}(0,1),
\label{eq:latent_deg_path}
\end{equation}
and the target velocity along this path is
\begin{equation}
u^{\star}(z_{\tau}, \tau)
= \frac{d z_{\tau}}{d \tau}
= z_{\mathrm{hr}} - \tilde{z}_{\mathrm{lr}}.
\label{eq:latent_deg_velocity}
\end{equation}
The refinement network $v_{\theta}$ is then trained with the flow-matching objective
\begin{equation}
\mathcal{L}_{\mathrm{FM}}
=
\mathbb{E}_{z_{\mathrm{hr}},\,\tilde{z}_{\mathrm{lr}},\,\tau}
\left[
\left\|
v_{\theta}(z_{\tau}, \tau \mid \tilde{z}_{\mathrm{lr}}, x_0, c)
-
\left(z_{\mathrm{hr}} - \tilde{z}_{\mathrm{lr}}\right)
\right\|_2^2
\right].
\label{eq:latent_deg_fm_loss}
\end{equation}

This design has two advantages.
First, pixel degradation preserves the coarse layout and motion cues required by the refinement stage.
Second, latent degradation weakens the one-to-one correspondence between degraded and target videos,
preventing the model from collapsing to a trivial upsampler.
Consequently, the refiner is encouraged to regenerate contact-sensitive structures,
such as the gripper boundary, object contours, and local interaction geometry,
which are essential for embodied manipulation.

At inference time, we do not synthesize a degraded target from the ground-truth video.
Instead, given the preview-stage output $z_{\mathrm{lr}}$,
we first upsample it to the high-resolution latent size and then inject mild latent perturbation:
\begin{equation}
\bar{z}_{\mathrm{lr}} = \mathrm{Up}(z_{\mathrm{lr}}),
\qquad
z^{0} = \alpha_{s_{\mathrm{inf}}}\,\bar{z}_{\mathrm{lr}} + \beta_{s_{\mathrm{inf}}}\,\epsilon,
\qquad \epsilon \sim \mathcal{N}(0, I),
\label{eq:latent_deg_inference}
\end{equation}
where $\mathrm{Up}(\cdot)$ denotes latent upsampling and $s_{\mathrm{inf}}$ is the latent degradation
strength used at inference.
Starting from $z^{0}$, the refinement model performs a small number of Euler updates:
\begin{equation}
z^{k+1}
=
z^{k}
+
\Delta \tau \,
v_{\theta}(z^{k}, \tau_k \mid z_{\mathrm{lr}}, x_0, c),
\qquad k=0,\ldots,K-1,
\label{eq:latent_deg_euler}
\end{equation}
and outputs the final refined latent $z^{K}$, which is decoded by the VAE decoder to obtain the high-resolution video.

In summary, latent degradation makes the second stage solve a \emph{detail regeneration} problem instead of a simple \emph{video super-resolution} problem, which is particularly beneficial in contact-rich robotic scenes where small structural errors in the end-effector region can lead to large semantic deviations in the predicted interaction process.

\section{Details of Dataset Preparation}
\subsection{Introduction of Datasets}
\noindent\textbf{LIBERO.} The LIBERO benchmark~\cite{liu2023libero} is a widely used robotic manipulation suite designed to test generalization across diverse tasks, object layouts, and language instructions. LIBERO contains long-horizon tabletop manipulation demonstrations with substantial variation in scene composition and target semantics, making it suitable for evaluating whether a video world model can both preserve interaction dynamics and follow task-level instructions. In our setting, each trajectory provides an initial observation, a language command, and the subsequent visual evolution of the manipulation process. 

\noindent\textbf{Real-World Dataset.} The real-world dataset is the Pine 7K dataset, a large-scale robotic manipulation dataset collected using Galaxea A1 robot platform. 
The dataset comprises \textbf{6,041 episodes} spanning \textbf{80 distinct manipulation tasks}, with a total of \textbf{505,779 frames} amounting to approximately \textbf{14 hours} of robot interaction data.
Each episode is recorded from two camera viewpoints: a \textit{scene} camera providing a third-person overview of the workspace, and a \textit{wrist}-mounted camera capturing the end-effector perspective.
All videos are encoded in H.264 at a resolution of $424 \times 240$ pixels and a frame rate of 10~FPS.
The task repertoire covers a diverse range of tabletop manipulation skills, including:
\begin{itemize}
    \item Pick-and-place (e.g., \texttt{pick\_lemon\_into\_the\_bowl})
    \item Stacking (e.g., \texttt{stack\_blue\_cup\_on\_the\_green\_cup})
    \item Pouring (e.g., \texttt{pour\_water\_into\_the\_cup});
    \item Multi-step compositions (e.g., \texttt{pick\_apple\_into\_bowl\_then\_watermelon\\\_and\_mango\_on\_plate});
    \item Other skills such as sweeping, sorting, and tool use (e.g., \texttt{sweep\_the\_fruits\\\_from\_left\_to\_right}).
\end{itemize}
The key statistics of the dataset are summarized in Table~\ref{tab:dataset_stats}.
\begin{table}[h]
    \centering
    \caption{Summary statistics of the Pine 7K dataset.}
    \label{tab:dataset_stats}
    \begin{tabular}{l r}
        \toprule
        \textbf{Statistic} & \textbf{Value} \\
        \midrule
        Robot platform          & Galaxea A1 \\
        Total tasks             & 80 \\
        Total episodes          & 6,041 \\
        Total frames            & 505,779 \\
        Total duration          & $\sim$14 hours \\
        Avg. episode length     & $\sim$84 frames (8.4s) \\
        Frame rate              & 10 FPS \\
        Resolution              & $424 \times 240$ \\
        Camera views            & 2 (scene + wrist) \\
        Video codec             & H.264 \\
        Action space            & 7-DoF (EE \& Joint) \\
        \bottomrule
    \end{tabular}
\end{table}
\subsection{Dataset Preprocessing}
For the LIBERO simulation dataset, we follow a preprocessing protocol similar to prior embodied world model works such as WorldVLA \cite{cen2025worldvla}. 
Starting from the official LIBERO demonstrations, we first replay each trajectory to obtain temporally consistent visual observations paired with the corresponding language instruction. 
We then discard invalid samples, including dummy episodes, corrupted rollouts, and trajectories whose visual observations are incomplete or inconsistent with the task execution. 
To standardize the training set, all retained trajectories are temporally aligned and resized to the target image resolution used by each stage of our model. 
After filtering, we split the processed dataset by task into training and test sets with a ratio of $8{:}2$, yielding around 5K valid trajectories in total for our experiments.

For the real-world dataset, we resize all videos to a unified resolution of $480 \times 720$.
To ensure a consistent prediction horizon for training and evaluation, we discard trajectories whose length is shorter than 49 frames.
The remaining videos are used to construct the real-world training and test sets for our experiments.

\subsection{Visualization of Real-World Dataset}
\label{sec:vis_realworld}
\begin{figure*}[t]
    \centering
    \includegraphics[width=\linewidth]{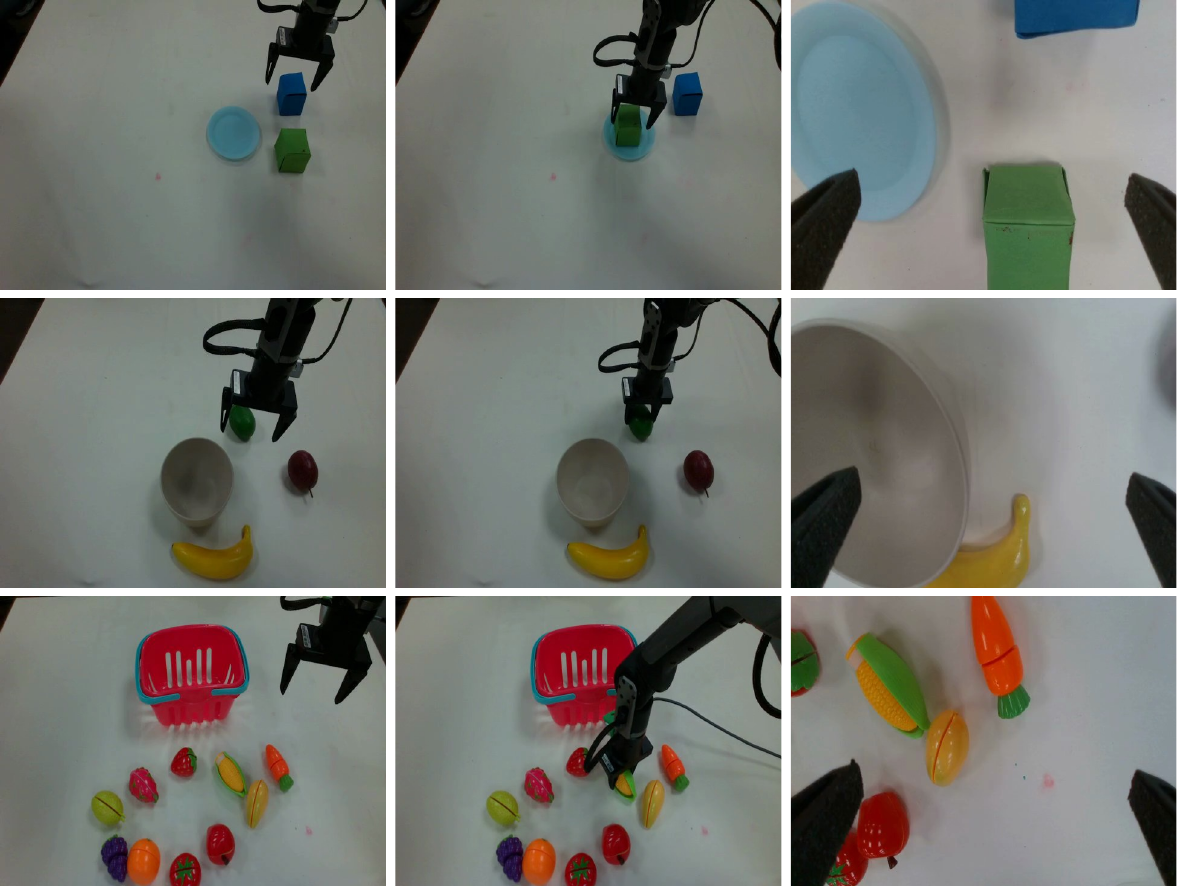}
    \caption{%
        Visualization of three randomly sampled tasks from the Pine 7K dataset.
        For each task, we show scene-camera frames (start, middle, end) and a wrist-camera frame from a representative episode.
        The scene view captures the global workspace context, while the wrist view provides a close-up of the end-effector interaction.
    }
    \label{fig:dataset_vis}
\end{figure*}

To provide an intuitive understanding of the training data, we visualize randomly sampled episodes from the Pine 7K dataset in Figure~\ref{fig:dataset_vis}.
We select three representative tasks spanning different manipulation primitives:
(1)~\texttt{pick\_green\_cube\_on\\\_plate\_and\_stack\_blue\_cube\_on\_green\_cube}, a multi-step task requiring sequential pick-place and stacking;
(2)~\texttt{pick\_lemon\_into\_the\_bowl}, a single-object pick-and-place task; and
(3)~\texttt{pick\_corn\_into\_basket}, a grasping task with a different target container.
For each task, we decompose a representative episode into temporal keyframes sampled from the scene camera (providing a third-person workspace overview) and the wrist-mounted camera (capturing the egocentric end-effector perspective).
The scene-camera sequence illustrates the full manipulation trajectory from the initial configuration to task completion, while the wrist view reveals fine-grained gripper--object interactions that are critical for contact-rich manipulation.
These visualizations highlight the diversity of the dataset in terms of object categories, manipulation skills, and visual appearance, demonstrating that the collected data covers a broad distribution of real-world tabletop scenarios.
\begin{table}[h]
\centering
\small
\setlength{\tabcolsep}{2pt}
\renewcommand{\arraystretch}{1.15}
\caption{Implementation details of DVG-WM.}
\label{tab:training_details}
\begin{tabular}{lll}
\toprule
\textbf{Module} & \textbf{Element} & \textbf{Detail} \\
\midrule

\multirow{5}{*}{System}
& GPU & 1$\times$ A100-80G (preview), 8$\times$ A100-80G (refinement) \\
& Optimizer & AdamW \\
& Video horizon & 49 frames \\
& HR resolution & $480 \times 720$ \\
& LR resolution & $256 \times 384$ \\
\midrule

\multirow{9}{*}{Preview stage}
& Backbone & CogVideoX-5B \\
& Training strategy & LoRA fine-tuning \\
& LoRA rank & 128 \\
& Tuned modules & Attention, FFN, LayerNorm \\
& Iterations & 10,000 \\
& Batch size & 4 \\
& GPU & 1$\times$ A100-80G \\
& Optimizer & AdamW \\
& Hyper-parameters & $\beta_1=0.9$, $\beta_2=0.95$, weight decay $=10^{-4}$,\\ & &grad clip $=0.1$ \\
\midrule

\multirow{7}{*}{Refinement stage}
& Backbone & CogVideoX-2B \\
& Objective & Flow matching \\
& Epochs & 10 \\
& Batch size & 6 \\
& GPU & 8$\times$ A100-80G \\
& Training time & $\sim$24 hours on LIBERO \\
& Hyper-parameters & Same as CogVideoX~\cite{yang2024cogvideox} \\
\midrule

\multirow{4}{*}{Action expert}
& Policy & Vision-only Diffusion Policy \\
& Input & Predicted frames at different timestamps \\
& Training data & $\sim$50 demonstrations per real-world task \\
& Planning horizon & 1 \\
\bottomrule
\end{tabular}
\end{table}
\begin{figure}[h]
    \centering
    \includegraphics[width=1\linewidth]{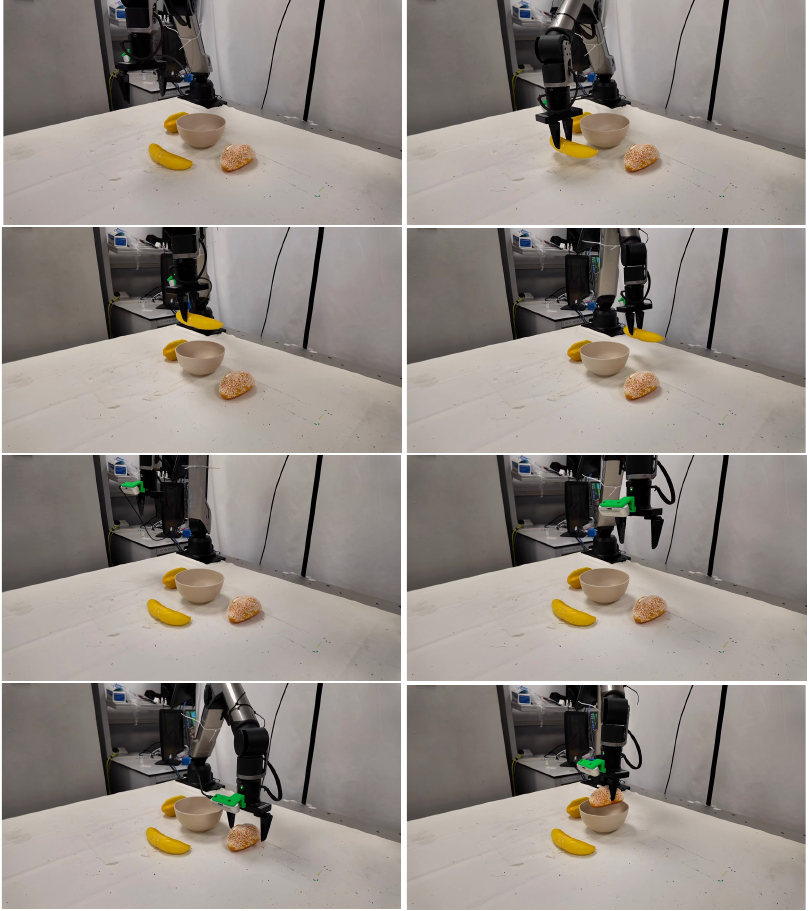}
    \caption{Examples of real-world demonstrations used in our experiments.
    We collect demonstrations for two manipulation tasks: moving banana and placing bread into the gray bowl.
    Each trajectory contains the initial observation, the language instruction, and the full visual process of the manipulation.}
    \label{fig:appendix_demo}
\end{figure}
\section{Training Details}

We summarize the implementation details of DVG-WM in \cref{tab:training_details} for reproducibility.
Unless otherwise specified, the training configurations follow the default settings of CogVideoX~\cite{yang2024cogvideox}.

\section{Details of Real-World Experiments}

We further evaluate DVG-WM on multiple real-world manipulation tasks to verify whether the proposed world model can support practical robotic planning beyond simulation.
The experiments are conducted on a Galaxea A1 robotic arm with a parallel gripper, and a UR7e arm using a RealSense L515 RGBD camera for visual observation.
The real-world dataset is collected under a fixed tabletop setup with layouts aligned to the training environment.

As shown in \cref{fig:appendix_demo}, we collect demonstrations for two representative pick-and-place style tasks:
\textit{moving banana} and \textit{placing bread into the gray bowl}.
These tasks are selected because they require the robot to accurately understand object identity, localize the target, and execute contact-rich interactions such as grasping, lifting, transporting, and placing.
Compared with simulation, the real-world setting introduces additional challenges including sensor noise, lighting variation, slight motion blur, and imperfect object poses, making it a more practical testbed for embodied world modeling.

For each task, around 50 human demonstrations are collected for imitation learning.
Each demonstration records the full manipulation trajectory from the initial scene to task completion, together with the corresponding language instruction.
Following the preprocessing protocol described in the main paper, all videos are resized to $480 \times 720$, and trajectories with fewer than 49 frames are discarded to maintain a consistent prediction horizon.
The retained demonstrations are then used to train both the world model and the action expert.

During evaluation, the world model predicts future visual trajectories conditioned on the initial observation and the task instruction, and the action expert converts the predicted futures into executable robot actions.
We report video quality metrics, Mask-IoU on embodiment regions, and the final task success rate to assess both visual prediction fidelity and control effectiveness.
This setup allows us to evaluate whether DVG-WM can generate realistic future interactions and provide useful planning signals for real-world robotic manipulation.
\FloatBarrier

\end{document}